\documentclass[acmtog]{acmart}
\usepackage{epsfig}
\usepackage{graphicx}
\usepackage{adjustbox}
\usepackage{amsmath}

\usepackage{amssymb}
\usepackage{environ}
\usepackage{tabularx}
\usepackage[table]{colortbl}
\usepackage[utf8]{inputenc}
\usepackage{color,soul}
\usepackage[font=small]{caption}
\usepackage{siunitx}

\NewEnviron{myequation}{%
    \begin{equation}
    \scalebox{1.5}{$\BODY$}
    \end{equation}
    }
    
\newcommand\norm[1]{\left\lVert#1\right\rVert}

\newcommand{\bp}{\textbf{p}}

\newcommand{\bo}{\textbf{o}}

\newcommand{\bc}{\textbf{c}}

\newcommand{\bz}{\textbf{z}}

\newcommand{\bq}{\textbf{q}}

\newcommand{\bx}{\textbf{x}}
\newcommand{\bg}{\textbf{g}}

\newcommand{\br}{\textbf{r}}
\newcommand{\bb}{\textbf{b}}
\newcommand{\bh}{\textbf{h}}

\newcommand{\etal}{\textit{et al}. }

\AtBeginDocument{%
  \providecommand\BibTeX{{%
    \normalfont B\kern-0.5em{\scshape i\kern-0.25em b}\kern-0.8em\TeX}}}

\setcopyright{acmlicensed}
\copyrightyear{2020}
\acmJournal{TOG}
\acmYear{2020}\acmVolume{39}\acmNumber{4}\acmArticle{60}\acmMonth{7} \acmDOI{10.1145/3386569.3392480}

\acmSubmissionID{553}

\begin{document}

\title{Robust Motion In-betweening}

\author{Félix G. Harvey}
\affiliation{%
  \institution{Polytechnique Montreal}
  \streetaddress{2500 Chemin de la Polytechique}
  \city{Montreal}
  \state{QC}
  \postcode{H3T 1J4}
  \country{Canada}
  }
\affiliation{%
  \institution{Mila}
  \streetaddress{6666 St-Urbain Street, \#200}
  \city{Montreal}
  \state{QC}
  \postcode{H2S 3H1}
  \country{Canada}
  }
\affiliation{%
  \institution{Ubisoft Montreal}
  \streetaddress{5505 Boul Saint-Laurent, \#2000}
  \city{Montreal}
  \state{QC}
  \postcode{H2T 1S6}
  \country{Canada}
  }
\email{felix.gingras-harvey@polymtl.ca}

\author{Mike Yurick}
\affiliation{%
  \institution{Ubisoft Montreal}
  \streetaddress{5505 Boul Saint-Laurent, \#2000}
  \city{Montreal}
  \state{QC}
  \postcode{H2T 1S6}
  \country{Canada}
  }
\email{mike.yurick@ubisoft.com}

\author{Derek Nowrouzezahrai}
\affiliation{%
  \institution{McGill University}
  \streetaddress{3480 University St}
  \city{Montreal}
  \state{QC}
  \postcode{H3A 0E9}
  \country{Canada}
  }
 \affiliation{%
  \institution{Mila}
  \streetaddress{6666 St-Urbain Street, \#200}
  \city{Montreal}
  \state{QC}
  \postcode{H2S 3H1}
  \country{Canada}
  }
\email{derek@cim.mcgill.ca}

\author{Christopher Pal}
  \affiliation{%
  \institution{CIFAR AI Chair}
  \streetaddress{661 University Ave., Suite 505}
  \city{Toronto}
  \state{ON}
  \postcode{M5G 1M1}
  \country{Canada}
  }
\affiliation{%
  \institution{Polytechnique Montreal}
  \streetaddress{2500 Chemin de la Polytechique}
  \city{Montreal}
  \state{QC}
  \postcode{H3T 1J4}
  \country{Canada}
  }
\affiliation{%
  \institution{Mila}
  \streetaddress{6666 St-Urbain Street, \#200}
  \city{Montreal}
  \state{QC}
  \postcode{H2S 3H1}
  \country{Canada}
  }
 \affiliation{%
  \institution{Element AI}
  \streetaddress{6650 St-Urbain Street, \#500}
  \city{Montreal}
  \state{QC}
  \postcode{H2S 3G9}
  \country{Canada}
  }

\renewcommand\shortauthors{Harvey \etal}
\begin{abstract}
    In this work we present a novel, robust transition generation technique that can serve as a new tool for 3D animators, based on adversarial recurrent neural networks. The system synthesizes high-quality motions that use temporally-sparse keyframes as animation constraints. This is reminiscent of the job of \textit{in-betweening} in traditional animation pipelines, in which an animator draws motion frames between provided keyframes. We first show that a state-of-the-art motion prediction model cannot be easily converted into a robust transition generator when only adding conditioning information about future keyframes. To solve this problem, we then propose two novel additive embedding modifiers that are applied at each timestep to latent representations encoded inside the network's architecture. One modifier is a \textit{time-to-arrival embedding} that allows variations of the transition length with a single model. The other is a \textit{scheduled target noise} vector that allows the system to be robust to target distortions and to sample different transitions given fixed keyframes. To qualitatively evaluate our method, we present a custom MotionBuilder plugin that uses our trained model to perform in-betweening in production scenarios. To quantitatively evaluate performance on transitions and generalizations to longer time horizons, we present well-defined in-betweening benchmarks on a subset of the widely used Human3.6M dataset and on LaFAN1, a novel high quality motion capture dataset that is more appropriate for transition generation. We are releasing this new dataset along with this work, with accompanying code for reproducing our baseline results.      
\end{abstract}



\begin{teaserfigure}
  \includegraphics[width=\textwidth]{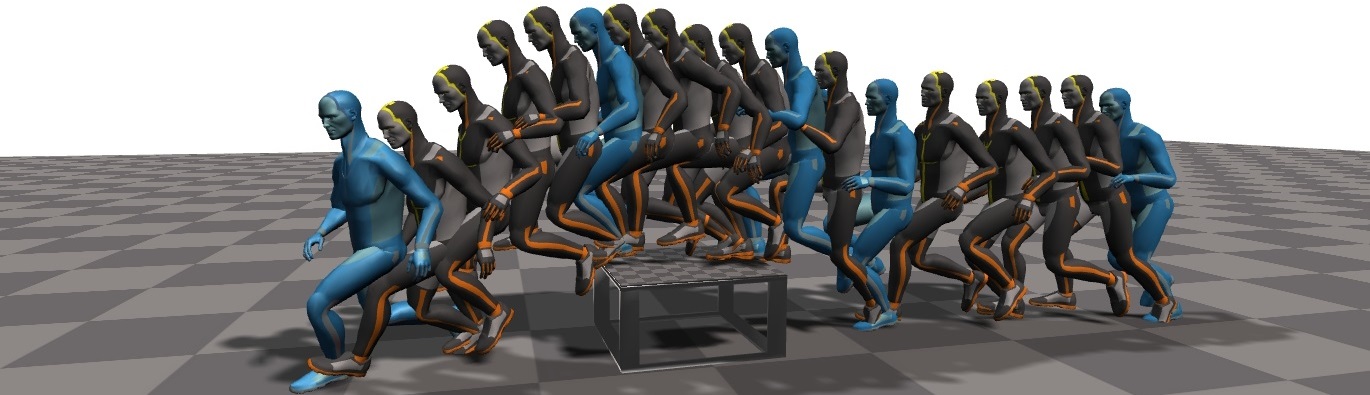}
  \caption{Transitions automatically generated by our system between target keyframes (in blue). For clarity, only one in four generated frames is shown. Our tool allows for generating transitions of variable lengths and for sampling different variations of motion given fixed keyframes.}
  \Description{Motion in-betweening example.}
  \label{fig:teaser}
\end{teaserfigure}

\maketitle

\section{Introduction}
Human motion is inherently complex and stochastic for long-term horizons. This is why Motion Capture (MOCAP) technologies still often surpass generative modeling or traditional animation techniques for 3D characters with many degrees of freedom. However, in modern video games, the number of motion clips needed to properly animate a complex character with rich behaviors is often very large and manually authoring animation sequences with keyframes or using a MOCAP pipeline are highly time-consuming processes. Some methods to improve curve fitting between keyframes \cite{Ciccone:2019:TOI:3306346.3322938} or to accelerate the MOCAP workflow \cite{holden2018robust} have been proposed to improve these processes. On another front, many auto-regressive deep learning methods that leverage high quality MOCAP for motion prediction have recently been proposed 
\cite{fragkiadaki2015recurrent, jain2016structural, martinez2017human, barsoum2018hp, pavllo2019modeling, gopalakrishnan2019neural, chiu2019action}. Inspired by these achievements, we build in this work a transition generation tool that leverages the power of Recurrent Neural Networks (RNN) as powerful motion predictors to go beyond keyframe interpolation techniques, which have limited expressiveness and applicability.

We start by building a state-of-the-art motion predictor based on several recent advances on modeling human motion with RNNs \cite{fragkiadaki2015recurrent, chiu2019action, pavllo2019modeling}. 
Using a recently proposed target-conditioning strategy \cite{harvey2018recurrent}, we convert this unconstrained predictor into a transition generator, and expose the limitations of such a conditioning strategy. 
These limitations include poor handling of transitions of different lengths for a single model, and the inherent determinism of the architectures. The goal of this work is to tackle such problems in order to present a new architecture that is usable in a production environment.

To do so, we propose two different additive modifiers applied to some of the latent representations encoded by the network. The first one is a \textit{time-to-arrival embedding} applied on the hidden representation of all inputs. This temporal embedding is similar to the positional encoding used in transformer networks \cite{vaswani2017attention} in natural language modeling, but serves here a different role. In our case, these embeddings evolve backwards in time from the target frame in order to allow the recurrent layer to have a continuous, dense representation of the number of timesteps remaining before the target keyframe must be reached. This proves to be essential to remove artifacts such as gaps or stalling at the end of transitions. The second embedding modifier is an additive \textit{scheduled target noise} vector that forces the recurrent layer to receive distorted target embeddings at the beginning of long transitions. The scheduled scaling reduces the norm of the noise during the synthesis in order to reach the correct keyframe. This forces the generator to be robust to noisy target embeddings. We show that it can also be used to enforce stochasticity in the generated transitions more efficiently than another noise-based method. We then further increase the quality of the generated transitions by operating in the Generative Adversarial Network (GAN) framework with two simple discriminators applied on different timescales.
%
%
%

This results in a temporally-aware, stochastic, adversarial architecture able to generate missing motions of variable length between sparse keyframes of animation. The network takes 10 frames of past context and a single target keyframe as inputs and produces a smooth motion that leads to the target, on time. It allows for cyclic and acyclic motions alike and can therefore help generate high-quality animations from sparser keyframes than what is usually allowed by curve-fitting techniques. Our model can fill gaps of an arbitrary number of frames under a soft upper-bound and we show that the particular form of temporal awareness we use is key to achieve this without needing any smoothing post-process. The resulting system allows us to perform robust, automatic in-betweening, or can be used to stitch different pieces of existing motions when blending is impossible or yields poor quality motion. 

Our system is tested in production scenarios by integrating a trained network in a custom plugin for Autodesk's MotionBuilder, a popular animation software, where it is used to greatly accelerate prototyping and authoring new animations.
In order to also quantitatively assess the performance of different methods on the transition generation task, we present the LaFAN1 dataset, a novel collection of high quality MOCAP sequences that is well-suited for transition generation. We define in-betweening benchmarks on this new dataset as well as on a subset of Human3.6M, commonly used in the motion prediction literature. Our procedure stays close to the common evaluation scheme used in many prediction papers and defined by Jain \etal \shortcite{jain2016structural}, but differs on some important aspects. First, we provide error metrics that take into consideration the global root transformation of the skeleton, which provides a better assessment of the absolute motion of the character in the world. This is mandatory in order to produce and evaluate valid transitions. Second, we train and evaluate the models in an action-agnostic fashion and report average errors on a large evaluation set, as opposed to the commonly used 8 sequences per action. We further report generalization results for transitions that are longer than those seen during training. Finally, we also report the Normalized Power Spectrum Similarity (NPSS) measure for all evaluations, as suggested by Gopalakrishnan \etal \shortcite{gopalakrishnan2019neural} which reportedly correlates better with human perception of quality.

Our main contributions can thus be summarized as follow:
\begin{itemize}
    \item Latent additive modifiers to convert state-of-the-art motion predictors into robust transition generators:
    \begin{itemize}
        \item A \textit{time-to-arrival embedding} allowing robustness to varying transition lengths,
        \item A \textit{scheduled target-noise} vector allowing variations in generated transitions,
    \end{itemize}
    \item New in-betweening benchmarks that take into account global displacements and generalization to longer sequences,
    \item LaFAN1, a novel high quality motion dataset well-suited for motion prediction that we make publicly available with accompanying code for reproducing our baseline results\footnote{\href{https://github.com/ubisoftinc/Ubisoft-LaForge-Animation-Dataset}{https://github.com/ubisoftinc/Ubisoft-LaForge-Animation-Dataset}}.
\end{itemize}

\section{Related Work}
\subsection{Motion Control}
We refer to motion control here as scenarios in which temporally-dense external signals, usually user-defined, are used to drive the generation of an animation. Even if the main application of the present work is not focused on online control, many works on motion control stay relevant to this research. Motion graphs \cite{arikan2002interactive, lee2002interactive, beaudoin2008motion, kovar2008motion} allow one to produce motions by traversing nodes and edges that map to character states or motions segments from a dataset. Safonova and Hodgins \cite{safonova2007construction} combine an interpolated motion graph to an anytime $A^*$ search algorithm in order produce transitions that respect some constraints. Motion matching \cite{ButtnerNuclai} is another search driven motion control technique, where the current character pose and trajectory are matched to segments of animation in a large dataset. Chai \& Hodgins, and Tautges \etal \shortcite{chai2005performance, tautges2011motion} rely on learning local PCA models on pose candidates from a motion dataset given low-dimensional control signals and previously synthesized poses in order to generate the next motion frame. All these techniques require a motion database to be loaded in memory or in the latter cases to perform searches and learning at run-time, limiting their scalability compared to generative models.

 Many machine learning techniques can mitigate these requirements. Important work has used the \textit{Maximum A Posteriori} (MAP) framework where a motion prior is used to regularize constraint(s)-related objectives to generate motion. \cite{chai2007constraint} use a statistical dynamics model as a motion prior and user constraints, such as keyframes, to generate motion. Min \etal \shortcite{min2009interactive} use deformable motion models and optimize the deformable parameters at run-time given the MAP framework. Other statistical models, such as Gaussian Processes \cite{min2012motion} and Gaussian Process Latent Variable Models \cite{grochow2004style, levine2012continuous, wang2008gaussian, ye2010synthesis} have been applied to the constrained motion control task, but are often limited by heavy run-time computations and memory requirements that still scale with the size of the motion database. As a result, these are often applied to separate types of motions and combined together with some post-process, limiting the expressiveness of the systems.

Deep neural networks can circumvent these limitations by allowing huge, heterogeneous datasets to be used for training, while having a fixed computation budget at run-time. Holden \etal \shortcite{holden2015learning, holden2016deep} use feed-forward convolutional neural networks to build a constrained animation synthesis framework that uses root trajectory or end-effectors' positions as control signals. Online control from a gamepad has also been tackled with phase-aware \cite{holden2017phase}, mode-aware \cite{zhang2018mode} and action-aware \cite{starke2019neural} neural networks that can automatically choose a mixture of network weights at run-time to disambiguate possible motions.
Recurrent Neural Networks (RNNs) on the other hand keep an internal memory state at each timestep that allows them to perform naturally such disambiguation, and are very well suited for modeling time series. Lee \etal \shortcite{lee2018interactive} train an RNN for interactive control using multiple control signals. These approaches \cite{holden2016deep, holden2017phase, zhang2018mode, lee2018interactive} rely on spatially or temporally dense signals to constrain the motion and thus reduce ambiguity. In our system, a character might have to precisely reach a temporally distant keyframe without any dense spatial or temporal information provided by the user during the transition. The spatial ambiguity is mostly alleviated by the RNN's memory and the target-conditioning, while the timing ambiguity is resolved in our case by time-to-arrival embeddings added to the RNN inputs. Remaining ambiguity can be alleviated with generative adversarial training \cite{goodfellow2014generative}, in which the motion generator learns to fool an additional discriminator network that tries to differentiate generated sequences from real sequences. 
Barsoum \etal \shortcite{barsoum2018hp} and Gui \etal \shortcite{gui2018adversarial} both design new loss functions for human motion prediction, while also using adversarial losses using different types of discriminators. These losses help reduce artifacts that may be produced by generators that average different modes of the plausible motions' distribution.

Motion control has also been addressed with Reinforcement Learning (RL) approaches, in which the problem is framed as a Markov Decision Process where \textit{actions} can correspond to actual motion clips \cite{lee2006precomputing, treuille2007near} or character states \cite{lee2010motion}, but again requiring the motion dataset to be loaded in memory at run-time. Physically-based control gets rid of this limitation by having the output of the system operate on a physically-driven character. Coros \etal \shortcite{coros2009robust} employ fitted value iteration with actions corresponding to optimized Proportional-Derivative (PD) controllers proposed by Yin \etal \shortcite{yin2007simbicon}. These RL methods operate on value functions that have discrete domains, which do not represent the continuous nature of motion and impose run-time estimations through interpolation. 

Deep RL methods, which use neural networks as powerful continuous function approximators have recently started being used to address these limitations. Peng \etal \shortcite{peng2017deeploco} apply a hierarchical actor-critic algorithm that outputs desired joint angles for PD-controllers. Their approach is applied on a simplified skeleton and does not express human-like quality of movement despite their style constraints. Imitation-learning based RL approaches \cite{ho2016generative, baram2016model} try to address this with adversarial learning, while others tackle the problem by penalizing distance of a generated state from a reference state \cite{peng2018deepmimic, bergamin2019drecon}. Actions as animation clips, or control fragments \cite{liu2017learning} can also be used in a deep-RL framework with Q-learning to drive physically-based characters. These methods show impressive results for characters having physical interactions with the world, while still being limited to specific skills or short cyclic motions. We operate in our case in the kinematics domain and train on significantly more heterogeneous motions. 
\subsection{Motion Prediction}
We limit here the definition of motion prediction to generating unconstrained motion continuation given single or multiple frames of animation as context. This task implies learning a powerful motion dynamics model which is useful for transition generation.
Neural networks have shown over the years to excel in such representation learning. Early work from Taylor \etal \shortcite{taylor2007modeling} using Conditional Restricted Boltzmann Machines showed promising results on motion generation by sampling at each timestep the next frame of motion conditioned on the current hidden state and $n$ previous frames. More recently, many RNN-based approaches have been proposed for motion prediction from a past-context of several frames, motivated by the representational power of RNNs for temporal dynamics. Fragkiadki \etal \shortcite{fragkiadaki2015recurrent} propose to separate spatial encoding and decoding from the temporal dependencies modeling with the Encoder-Recurrent-Decoder (ERD) networks, while Jain \etal \shortcite{jain2016structural} apply structural RNNs to model human motion sequences represented as spatio-temporal graphs. 
Other recent approaches \cite{martinez2017human, tang2018long, chiu2019action, gopalakrishnan2019neural, liu2019towards, pavllo2019modeling} investigate new architectures and loss functions to further improve short-term and long-term prediction of human motion. Others \cite{li2017auto, ghosh2017learning} investigate ways to prevent divergence or collapsing to the average pose for long-term predictions with RNNs. 
In this work, we start by building a powerful motion predictor based on the state-of-the-art recurrent architecture for long-term prediction proposed by Chiu \etal \shortcite{chiu2019action}.
We combine this architecture with the feed-forward encoders of Harvey \etal \shortcite{harvey2018recurrent} applied to different parts of the input to allow our embedding modifiers to be applied on distinct parts of the inputs. In our case, we operate on joint-local quaternions for all bones, except for the root, for which we use quaternions and translations local to the last seed frame.  
%
\subsection{Transition generation}
We define transition generation as a type of control with temporally sparse spatial constraints, i.e. where large gaps of motion must be filled without explicit conditioning during the missing frames such as trajectory or contact information. This is related to keyframe or motion interpolation (e.g. \cite{Ciccone:2019:TOI:3306346.3322938}), but our work extends interpolation in that the system allows for generating whole cycles of motion, which cannot be done by most key-based interpolation techniques, such a spline fitting. Pioneering approaches \cite{Witkin:1988:SC:54852.378507, cohen1996efficient} on transition generation and interpolation used spacetime constraints and inverse kinematics to produce physically-plausible motion between keyframes. Work with probabilistic models of human motion have also been used for filling gaps of animation. These include the MAP optimizers of Chai \etal \shortcite{chai2007constraint} and Min \etal \shortcite{min2009interactive}, the Gaussian process dynamical models from Wang \etal \shortcite{wang2008gaussian} and Markov models with dynamic auto-regressive forests from Lehrmann \etal \shortcite{lehrmann2014efficient}. All of these present specific models for given action and actors. This can make combinations of actions look scripted and sequential. The scalability and expressiveness of deep neural networks has been applied to keyframe animation by Zhang \etal \shortcite{zhang2018data}, who use an RNN conditioned on key-frames to produce jumping motions for a simple 2D model. Harvey \etal \shortcite{harvey2018recurrent} present Recurrent Transition Networks (RTN) that operate on a more complex human character, but work on fixed-lengths transitions with positional data only and are deterministic. We use the core architecture of the RTN as we make use of the separately encoded inputs to apply our latent modifiers.
Hernandez \etal \shortcite{hernandez2019human} recently applied convolutional adversarial networks to pose the problem of prediction or transition generation as an in-painting one, given the success of convolutional generative adversarial networks on such tasks. They also propose frequency-based losses to assess motion quality, but do not provide a detailed evaluation for the task of in-betweening.
%
%
%
%
%
\section{Methods}\label{section:model}
\subsection{Data formatting}\label{section:data}
We use a humanoid skeleton that has $j = 28$ joints when using the Human3.6M dataset and $j=22$ in the case of the LaFAN1 dataset. We use a local quaternion vector $\bq_t$ of $j* 4$ dimensions as our main data representation along with a 3-dimensional global root velocity vector $\dot{\br}_t$ at each timestep $t$. 
%
We also extract from the data, based on toes and feet velocities, contact information as a binary vector $\bc_t$ of $4$ dimensions that we use when working with the LaFAN1 dataset.
The offset vectors $\bo_t^r$ and $\bo_t^q$ contain respectively the global root position's offset and local-quaternions' offsets from the target keyframe at time $t$. Even though the quaternion offset could be expressed as a valid, normalized quaternion, we found that using simpler element-wise linear differences simplifies learning and yields better performance. When computing our positional loss, we reformat the predicted state into a global positions vector $\bp_{t+1}$ using $\bq_{t+1}$, $\br_{t+1}$ and the stored, constant local bone translations $\bb$ by performing Forward Kinematics (FK). The resulting vector $\bp_{t+1}$ has $j * 3$ dimensions. We also retrieve through FK the global quaternions vector $\bg_{t+1}$, which we use for quantitatively evaluating transitions.
%
%
%
%

The discriminator use as input sequences of 3-dimensional vectors of global root velocities $\dot\br$, concatenated with $\bx$ and $\dot\bx$, the root-relative positions and velocities of all other bones respectively. The vectors $\bx$ and $\dot\bx$ both have $(j-1)*3$ dimensions.

To simplify the learning process, we rotate each input sequence seen by the network around the $Y$ axis (up) so that the root of the skeleton points towards the $X^+$ axis on the last frame of past context. Each transition thus starts with the same global horizontal facing. We refer to rotations and positions relative to this frame as \textit{global} in the rest of this work. When using the network inside a content creation software, we store the applied rotation in order to rotate back the generated motion to fit the context. Note however that this has no effect on the public dataset Human3.6M since root transformations are set to the identity on the first frame of any sequences, regardless of the actual world orientation. We also augment the data by mirroring the sequences over the $X^+$ axes with a probability of $0.5$ during training.

%
%
\subsection{Transition Generator}\label{section:overview}
\begin{figure}[h]
\begin{center}
\includegraphics[width=0.45\textwidth]{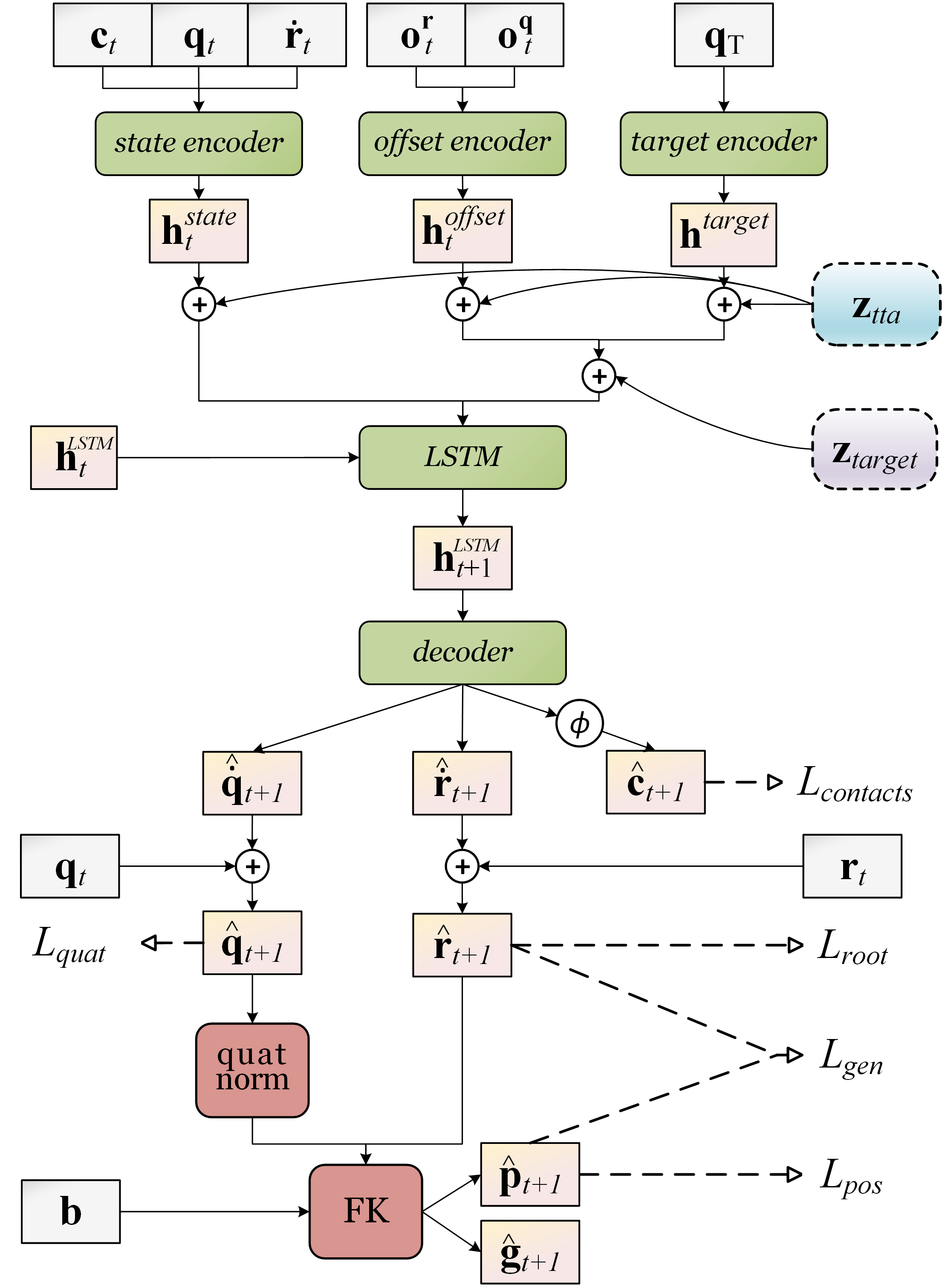}
  \caption{\textbf{Overview of the $\mathrm{\textbf{TG}}_{\mathit{complete}}$ architecture for in-betweening.} Computations for a single timestep are shown. Visual concatenation of input boxes or arrows represents vector concatenation. Green boxes are the jointly trained neural networks. Dashed boxes represent our two proposed embedding modifiers. The "quat norm" and "FK" red boxes represent the quaternion normalization and Forward Kinematics operations respectively. The $\oplus$ sign represents element-wise addition and $\phi$ is the sigmoid non-linearity. Outputs are linked to associated losses with dashed lines.}
  \label{fig:overview}
\end{center}
\vspace{-0.5cm}
\end{figure}

Figure \ref{fig:overview} presents a visual depiction of our recurrent generator for a single timestep. It uses the same input separation used by the RTN network \cite{harvey2018recurrent}, but operates on angular data and uses FK in order to retrieve global positions \cite{pavllo2019modeling}. It is also augmented with our latent space modifiers $\bz_{\mathit{tta}}$ and $\bz_{\mathit{target}}$. Finally it also uses different losses, such as an adversarial loss for improved realism of the generated motions.

As seen in Figure \ref{fig:overview}, the generator has three different encoders that take the different data vectors described above as inputs; the character state encoder, the offset encoder, and the target encoder. The encoders are all fully-connected Feed-Forward Networks (FFN) with a hidden layer of 512 units and an output layer of 256 units. All layers use the Piecewise Linear Activation function (PLU) \cite{nicolae2018plu}, which performed slightly better than Rectified Linear Units (ReLU) in our experiments. The time-to-arrival embedding $\bz_{tta}$ has 256 dimensions and is added to the latent input representations. 
Offset and target embeddings $\bh_t^{\mathit{offset}}$ and $\bh_t^{\mathit{target}}$ are then concatenated and added to the 512-dimensional target-noise vector $\bz_{\mathit{target}}$. 
Next, the three augmented embeddings are concatenated and fed as input to a recurrent Long-Short-Term-Memory (LSTM) layer. The embedding from the recurrent layer, $\bh_t^{\mathit{LSTM}}$ is then fed to the decoder, another FFN with two PLU hidden layers of 512 and 256 units respectively and a linear output layer. The resulting output is separated into local-quaternion and root velocities $\hat{\dot{\bq}}_{t+1}$ and $\hat{\dot{\br}}_{t+1}$ to retrieve the next character state. When working with the LaFAN1 dataset, the decoder has four extra output dimensions that go through a sigmoid non-linearity $\phi$ to retrieve contact predictions $\hat{\bc}_{t+1}$.
%
%
The estimated quaternions $\hat{\bq}_{t+1}$ are normalized as valid unit quaternions and used along with the new root position $\hat{\br}_{t+1}$ and the constant bone offsets $\bb$ to perform FK and retrieve the new global positions $\hat{\bp}_{t+1}$.

\subsection{Time-to-arrival embeddings}\label{sec:temporal}
We present here our method to allow robustness to variable lengths of in-betweening. In order to achieve this, simply adding conditioning information about the target keyframe is insufficient since the recurrent layer must be aware of the number of frames left until the target must be reached. This is essential to produce a smooth transition without teleportation or stalling. Transformer networks \cite{vaswani2017attention} are attention-based models that are increasingly used in natural language processing due to their state-of-the-art modeling capacity. They are sequence-to-sequence models that do not use recurrent layers but require positional encodings that modify a word embedding to represent its location in a sentence. Our problem is also a sequence-to-sequence task where we translate a sequence of seed frames to a transition sequence, with additional conditioning on the target keyframe. Although our generator does use a recurrent layer, it needs time-to-arrival awareness in order to gracefully handle transitions of variable lengths. To this end, we use the mathematical formulation of positional encodings, that we base in our case on the time-to-arrival to the target:
\vskip -0.25cm
 \begin{table}[h]
     \begin{adjustbox}{width=\columnwidth,center}
         \begin{tabular}{c c}
                \vbox{\begin{equation} \Huge%
                \mbox{\fontsize{23}{0}\selectfont\(
                \bz_{tta, 2i} = sin \left(\frac{tta}{\mathit{basis}^{2i / d}}\right)%
                \)} %
                \end{equation}}
                 & \vbox{\begin{equation}\Huge%
                 \mbox{\fontsize{23}{0}\selectfont\(%
                 \bz_{tta, 2i+1} = cos\left(\frac{tta}{\mathit{basis}^{2i / d}}\right)%
                 \)}%
                 \end{equation}}
         \end{tabular}
     \end{adjustbox}
 \end{table}
\vskip -0.25cm
where $\mathit{tta}$ is the number of timesteps until arrival and the second subscript of the vector $\bz_{\mathit{tta}, \_}$ represents the dimension index. The value $d$ is the dimensionality of the input embeddings, and $i \in [0, ..., d/2]$. The $\mathit{basis}$ component influences the rate of change in frequencies along the embedding dimensions. It is set to $\num{10000}$ as in most transformer implementations.

Time-to-arrival embeddings thus provide continuous codes that will shift input representations in the latent space smoothly and uniquely for each transition step due to the phase and frequency shifts of the sinusoidal waves on each dimension. Such embedding is thus bounded, smooth and dense, three characteristics beneficial for learning. Its additive nature makes it harder for a neural network to ignore, as can be the case with concatenation methods. This follows the successful trend in computer vision \cite{dumoulin2017learned, perez2018film} of conditioning through transformations of the latent space instead of conditioning with input concatenation.
%
%
%
In these cases, the conditioning signals are significantly more complex and the affine transformations need to be learned, whereas Vaswani \etal \shortcite{vaswani2017attention} report similar performance when using this sine-based formulation as when using learned embeddings.

It is said that positional encodings can generalize to longer sequences in the natural language domain. However, since $\bz_{\mathit{tta}}$ evolves backwards in time to retrieve a time-to-arrival representation, generalizing to longer sequences becomes a more difficult challenge. Indeed, in the cases of Transformers (without temporal reversal), the first embeddings of the sequence are always the same and smoothly evolve towards new ones when generalizing to longer sequences. In our case, longer sequences change drastically the initial embedding seen and may thus generate unstable hidden states inside the recurrent layer before the transition begins. This can hurt performance on the first frames of transitions when extending the time-horizon after training. To alleviate this problem, we define a maximum duration in which we allow $\bz_{\mathit{tta}}$ to vary, and fix it past this maximum duration. Precisely, the maximum duration $T_{\mathit{max}}(\bz_{\mathit{tta}})$ is set to $T_{max}(\mathit{trans}) + T_{past} - 5$, where $T_{\mathit{max}}(\mathit{trans})$ is the maximum transition length seen during training and $T_{\mathit{past}}$ is the number of seed frames given before the transition. This means that when dealing with transitions of length $T_{\mathit{max}}(\mathit{trans})$, the model sees a constant $\bz_{\mathit{tta}}$ for 5 frames before it starts to vary.
This allows the network to handle a constant $\bz_{\mathit{tta}}$ and to keep the benefits of this augmentation even when generalizing to longer transitions. Visual representations of $\bz_{\mathit{tta}}$ and the effects $T_{\mathit{max}}(\bz_{\mathit{tta}})$ are shown in Appendix \ref{app:tta}.

\begin{figure}[h]
\begin{center}
\centerline{\includegraphics[width=0.49\textwidth]{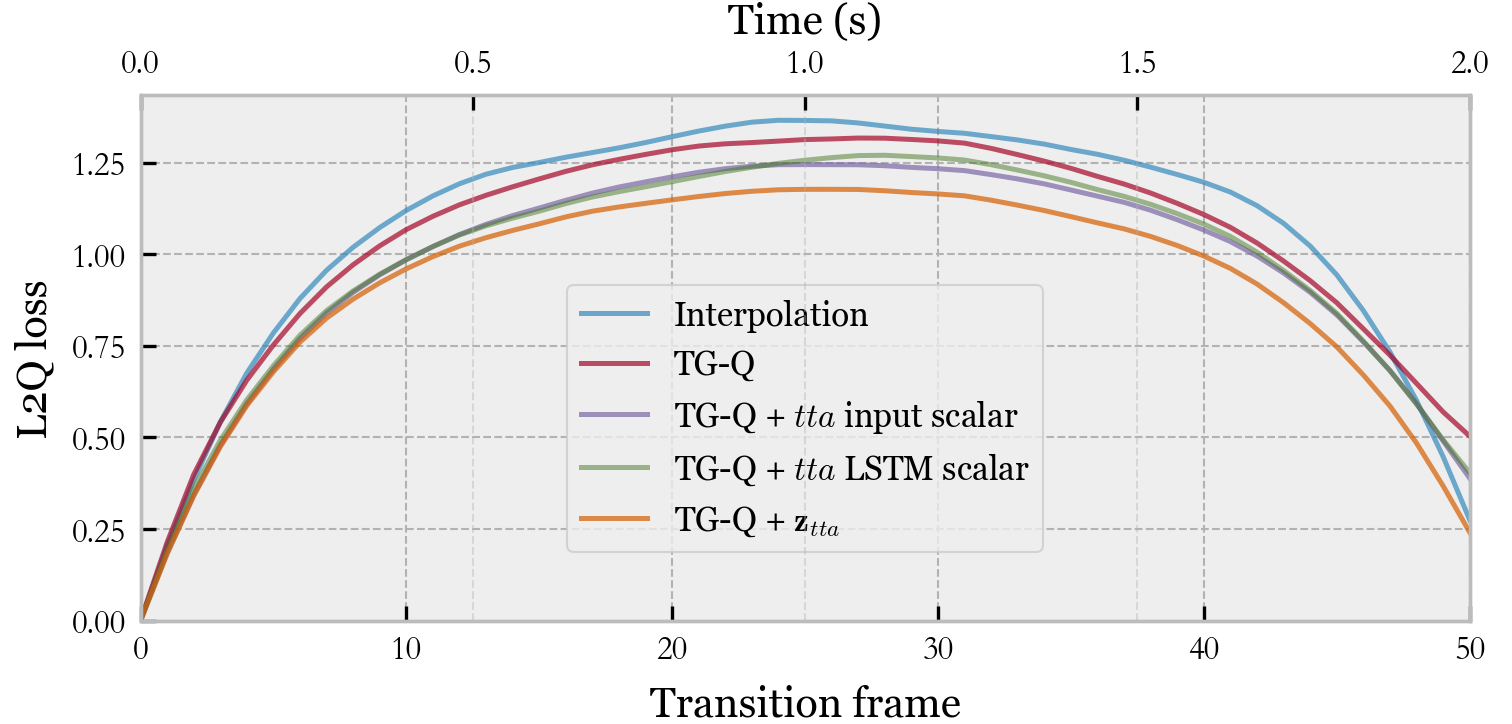}}
  \caption{\textbf{Reducing the L2Q loss with $\bz_{\mathit{tta}}$}. We compare simple interpolation with our temporally unaware model (TG-Q) on the walking subset of Human 3.6M. We further test two strategies based on adding a single $tta$ dimension either to the character state (TG-Q + $tta$ input scalar) or the LSTM inputs (TG-Q + $tta$ LSTM scalar). Finally, our use of time-to-arrival embeddings (TG-Q + $\bz_{\mathit{tta}}$) yields the best results, mostly noticeable at the end of transitions, where the generated motion is smoother than interpolation.}
  \label{fig:tta_effects}
\end{center}
\vspace{-0.5cm}
\end{figure}

We explored simpler approaches to induce temporal awareness, such as concatenating a time-to-arrival dimension either to the inputs of the state encoder, or to the LSTM layer's inputs. This $tta$ dimension is a single scalar increasing from 0 to 1 during the transition. Its period of increase is set to $T_{\mathit{max}}(\bz_{\mathit{tta}})$. Results comparing these methods with a temporally unaware network, and our use of $\bz_{\mathit{tta}}$ can be visualized in Figure \ref{fig:tta_effects}.

\subsection{Scheduled target noise}\label{sec:targetnoise}
\begin{figure}[h]
\begin{center}
\centerline{\includegraphics[width=0.45\textwidth]{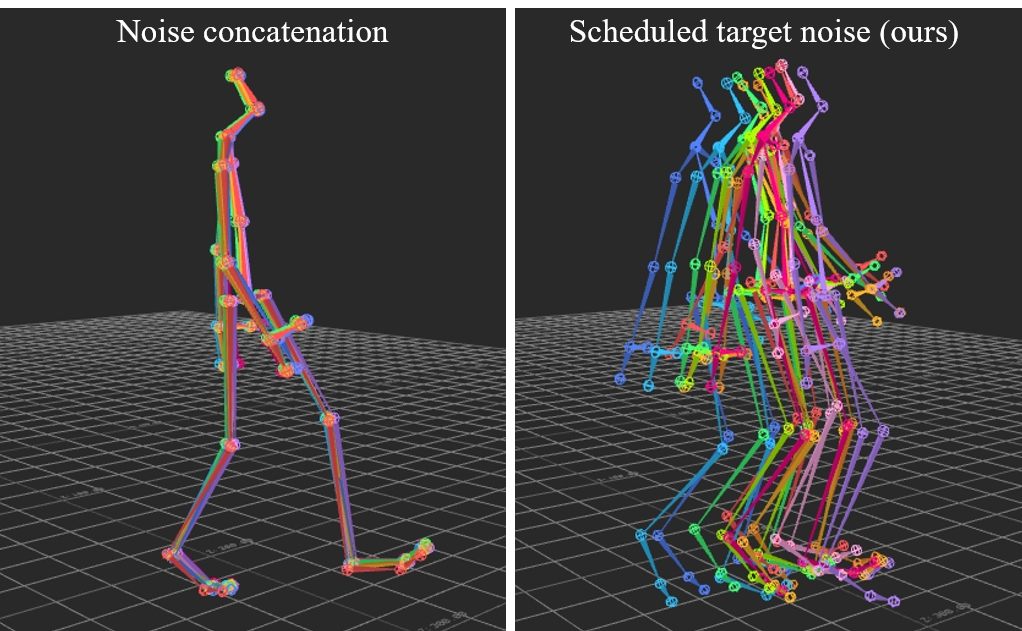}}
  \caption{\textbf{Increasing variability with $\bz_{\mathit{target}}$}. We compare $\bz_{\mathit{concat}}$ (left) against $\bz_{\mathit{target}}$ (right) midway in a 100-frames transition re-sampled 10 times. The generator successfully learns to ignore $\bz_{\mathit{concat}}$ while $\bz_{\mathit{target}}$ is imposed and leads to noticeable variations with controllable scale.}
  \label{fig:noisecomp}
\end{center}
\vspace{-0.5cm}
\end{figure}

Another contribution of this work is to improve robustness to keyframe modifications and to enforce diversity in the generated transitions given a fixed context. To do so we propose a scheduled target-distortion strategy.  We first concatenate the encoded embeddings $\bh_t^\mathit{offset}$ and $\bh_t^\mathit{target}$ of the current offset vector and the target keyframe respectively. We then add to the resulting vector the target noise vector $\bz_{\mathit{target}}$, sampled once per sequence from a spherical, zero-centered Gaussian distribution $\mathcal{N}(0, I * \sigma_{\mathit{target}})$. The standard deviation $\sigma_{\mathit{target}}$ is an hyper-parameter controlling the level of accepted distortion. In order to produce smooth transitions to the target, we then define a target noise multiplier $\lambda_{\mathit{target}}$, responsible for scaling down $\bz_{\mathit{target}}$ as the number of remaining timesteps goes down. We define a period of noise-free generation (5 frames) where $\lambda_{\mathit{target}}=0$ and a period of linear decrease of the target-noise (25 frames) to produce our noise-scale schedule. Beyond 30 frames before the target, the target-noise is therefore constant and $\lambda_{\mathit{target}} = 1$. 
\begin{align}
    \lambda_{\mathit{target}}= 
        \begin{cases}
            1                    & \text{if } \mathit{tta} \geq 30\\
            \frac{\mathit{tta - 5}}{25}   & \text{if } 5 \leq \mathit{tta} < 30\\
            0                   & \text{if } \mathit{tta} < 5\\
        \end{cases}
\end{align}
Since this modifier is additive, it also corrupts time-to-arrival information, effectively distorting the timing information. This allows to modify the pace of the generated motion. Our target noise schedule is intuitively similar to an agent receiving a distorted view of its long-term target, with this goal becoming clearer as the agent advances towards it. This additive embedding modifier outperformed in our experiments another common approach in terms of diversity of the transitions while keeping the motion plausible. Indeed a widespread approach to conditional GANs is to use a noise vector $\bz_{\mathit{concat}}$ as additional input to the conditioned generator in order to enable stochasticity and potentially disambiguate the possible outcomes from the condition (e.g. avoid mode collapsing). However in highly constrained cases like ours, the condition is often informative enough to obtain good performance, especially at the beginning of the training. This leads to the generator learning to ignore the additional noise, as observed in our tests (see Figure \ref{fig:noisecomp}). We thus force the transition generator to be stochastic by using $\bz_{\mathit{target}}$ to distort its view of the target and current offsets. 
\subsection{Motion Discriminators}
A common problem with reconstruction-based losses and RNNs is the \textit{blurriness} of the results, which is translated into collapse to the average motion and foot slides when predicting motion. The target keyframe conditioning can slightly alleviate these problems, but additional improvement comes from our use of adversarial losses, given by two discriminator networks. We use two variants of a relatively simple feed-forward architecture for our discriminators, or \textit{critics}, $\mathcal{C}_1$ and $\mathcal{C}_2$. Each discriminator has 3 fully-connected layers, with the last one being a 1D linear output layer. $\mathcal{C}_1$ is a \textit{long-term} critic that looks at sliding windows of 10 consecutive frames of motion and $\mathcal{C}_2$ is the \textit{short-term} critic and looks at windows of instant motion over 2 frames. Both critics have 512 and 256 units in their first and second hidden layers respectively. The hidden layers use ReLU activations. We average the discriminator scores over time in order to produce a single scalar loss. A visual summary of our sliding critics is presented in Appendix \ref{app:critics}.
%
%
%
%
\subsection{Losses}\label{sec:losses}
In order to make the training stable and to obtain the most realistic results, we use multiple loss functions as complementary soft constraints that the neural network learns to respect.

%

\subsubsection{Reconstruction Losses}
All of our reconstruction losses for a predicted sequence $\hat{X}$ given its ground-truth $X$ are computed with the L1 norm:
%
 \begin{table}[H]
     \begin{adjustbox}{width=\columnwidth,center}
         \begin{tabular}{c c}
                 \vbox{\begin{equation} \Huge L_{\mathit{quat}} = \frac{1}{T} \sum^{T-1}_{t=0} \|\hat{\bq}_{t} - \bq_{t}\|_1 \end{equation}}
                 & \vbox{\begin{equation} \Huge L_{\mathit{root}} = \frac{1}{T} \sum^{T-1}_{t=0} \|\hat{\br}_{t} - \br_{t}\|_1 \end{equation}} \\
                
                 \vbox{\begin{equation} \Huge L_{\mathit{pos}} = \frac{1}{T} \sum^{T-1}_{t=0} \|\hat{\bp}_{t} - \bp_{t}\|_1 \end{equation}}
                 & \vbox{\begin{equation} \Huge L_{\mathit{contacts}} = \frac{1}{T} \sum^{T-1}_{t=0} \|\hat{\bc}_{t} - \bc_{t}\|_1 \end{equation}}
         \end{tabular}
     \end{adjustbox}
     \vskip -0.25cm
 \end{table}
where $T$ is the sequence length. The two main losses that we use are the local-quaternion loss $L_{\mathit{quat}}$ and the root-position loss $L_{\mathit{root}}$. The former is computed over all joints' local rotations, including the root-node, which in this case also determines global orientation. The latter is responsible for the learning of the global root displacement. As an additional reconstruction loss, we use a positional-loss $L_{\mathit{pos}}$ that is computed on the global position of each joints retrieved through FK. In theory, the use of $L_{\mathit{pos}}$ isn't necessary to achieve a perfect reconstruction of the character state when using $L_{\mathit{quat}}$ and $L_{\mathit{root}}$, but as noted by Pavllo \etal \shortcite{pavllo2019modeling}, using global positions helps to implicitly weight the orientation of the bone's hierarchy for better results. As we will show in Section \ref{sec:human_transition}, adding this loss indeed improves results on both quaternion and translation reconstructions. Finally, in order to allow for runtime Inverse-Kinematics correction (IK) of the legs inside an animation software, we also use a contact prediction loss $L_{\mathit{contacts}}$, between predicted contacts $\hat{\bc}_t$ and true contacts $\bc_t$. We use the contact predictions at runtime to indicate when to perform IK on each leg. This loss is used only for models trained on the LaFAN1 dataset and that are deployed in our MotionBuilder plugin. 

\subsubsection{Adversarial Losses}
We use the Least Square GAN (LSGAN) formulation \cite{mao2017least}. As our discriminators operate on sliding windows of motion, we average their losses over time. Our LSGAN losses are defined as follows:
\begin{align}
     L_{\mathit{gen}} &= \frac{1}{2} \mathbb{E}_{\mathrm{X}_{\mathit{p}}, \mathrm{X}_{\mathit{f}} \sim p_{Data}}[(D({\mathrm{X}}_{\mathit{p}}, G(X_{\mathit{p}}, \mathrm{X}_{\mathit{f}}), {\mathrm{X}}_{\mathit{f}}) - 1)^2], \\
     L_{\mathit{disc}} &= \frac{1}{2} \mathbb{E}_{\mathrm{X}_{\mathit{p}}, \mathrm{X}_{\mathit{trans}}, \mathrm{X}_{\mathit{f}} \sim p_{Data}}[(D({\mathrm{X}}_{\mathit{p}}, \mathrm{X}_{\mathit{trans}}, {\mathrm{X}}_{\mathit{f}}) -1)^2] \nonumber\\
                        &+ \frac{1}{2} \mathbb{E}_{\mathrm{X}_{\mathit{p}}, \mathrm{X}_{\mathit{f}} \sim p_{Data}}[(D({\mathrm{X}}_{\mathit{p}}, G(X_{\mathit{p}}, \mathrm{X}_{\mathit{f}}),{\mathrm{X}}_{\mathit{f}}))^2],
 \end{align}
where $\mathrm{X}_{\mathit{p}}$, $\mathrm{X}_{\mathit{f}}$, and $\mathrm{X}_{\mathit{trans}}$ represent the past context, target state, and transition respectively, in the discriminator input format described in Section \ref{section:data}. $G$ is the transition generator network. 
Both discriminators use the same loss, with different input sequence lengths. 
%

%
%
\subsection{Training}

\subsubsection{Progressive growing of transitions}
In order to accelerate training, we adopt a curriculum learning strategy with respect to the transition lengths. Each training starts at the first epoch with $P_{\mathit{min}} = \tilde{P}_{\mathit{max}} = 5$, where $P_{\mathit{min}}$ and $\tilde{P}_{\mathit{max}}$ are the minimal and current maximal transition lengths. During training, we increase $\tilde{P}_{\mathit{max}}$ until it reaches the true maximum transition length $P_{\mathit{max}}$. The increase rate is set by number of epochs $n_{\mathit{ep-max}}$ by which we wish to have reached $\tilde{P}_{\mathit{max}} = P_{\mathit{max}}$. For each minibatch, we sample uniformly the current transition length between $P_{\mathit{min}}$ and $\tilde{P}_{\mathit{max}}$, making the network train with variable length transitions, while beginning the training with simple tasks only. In our experiments, this leads to similar results as using any teacher forcing strategy, while accelerating the beginning of training due to the shorter batches. Empirically, it also outperformed gradient clipping. At evaluation time, the transition length is fixed to the desired length.

\subsubsection{Sliding critics}
In practice, our discriminators are implemented as 1D temporal convolutions, with strides of 1, without padding, and with receptive fields of 1 in the last 2 layers, yielding parallel feed-forward networks for each motion window in the sequence.

\subsubsection{Hyperparameters} 
In all of our experiments, we use minibatches of 32 sequences of variable lengths as explained above. We use the AMSgrad optimizer \cite{sashank2018amsgrad} with a learning rate of 0.001 and adjusted parameters ($\beta_1 = 0.5, \beta2 = 0.9$) for increased stability. We scale all of our losses to be approximately equal on the LaFAN1 dataset for an untrained network before tuning them with custom weights. In all of our experiments, these relative weights (when applicable) are of 1.0 for $L_{\mathit{quat}}$ and $L_{\mathit{root}}$, 0.5 for $L_{\mathit{pos}}$, and 0.1 for $L_{\mathit{gen}}$ and $L_{\mathit{contacts}}$. The target noise's standard deviation $\sigma_{\mathit{target}}$ is 0.5. In experiments on Human3.6M, we set $n_{\mathit{ep-max}}$ to 5 while it is set to 3 on the larger LaFAN1 dataset. 

\section{Experiments and Results} \label{sec:results}
\subsection{Motion prediction}\label{sec:human_prediction}
\begin{table*}[ht]
    \centering
    \caption{\textbf{Unconstrained motion prediction results on Human 3.6M}. The VGRU-d/rl models are from \cite{gopalakrishnan2019neural}. The TP-RNN is from \cite{chiu2019action} and has to our knowledge the best published results on motion prediction for this benchmark. Our model, ERD-QV is competitive with the state-of-the-art on angular errors and improves performance with respect to the recently proposed NPSS metric on all actions.}\label{table:prediction}
    \footnotesize
    \setlength\tabcolsep{1.2pt}
    \begin{tabular}{lcccccccccccccccccccccccccccc}
        &\multicolumn{7}{c}{\textbf{Walking}} &\multicolumn{7}{c}{\textbf{Eating}} &\multicolumn{7}{c}{\textbf{Smoking}} &\multicolumn{7}{c}{\textbf{Discussion}}\\
        &\multicolumn{6}{c}{MAE} &\multicolumn{1}{c}{NPSS} &\multicolumn{6}{c}{MAE} &\multicolumn{1}{c}{NPSS} &\multicolumn{6}{c}{MAE} &\multicolumn{1}{c}{NPSS} &\multicolumn{6}{c}{MAE} &\multicolumn{1}{c}{NPSS} \\
        \cmidrule(l){2-7} \cmidrule(lr){8-8} \cmidrule(l){9-14} \cmidrule(lr){15-15} \cmidrule(l){16-21} \cmidrule(lr){22-22} \cmidrule(l){23-28} \cmidrule(lr){29-29} 
        milliseconds &80&160&320&500&560&1000 &0-1000  &80&160&320&500&560&1000 &0-1000 &80&160&320&500&560&1000 &0-1000 &80&160&320&500&560&1000 &0-1000\\
        \cmidrule{1-8}\cmidrule(l){9-15} \cmidrule(l){16-22}\cmidrule(l){23-29}
        Zero-Vel    &0.39	&0.68	&0.99	&1.15	&1.35	&1.32	&0.1418 &0.27	&0.48	&0.73	&0.86	&1.04	&1.38	&0.0839 &0.26	&0.48	&0.97	&0.95	&1.02	&1.69	&0.0572 	&0.31 &0.67	&0.94	&1.04	&1.41	&1.96	&0.1221\\
        \cmidrule{1-8}\cmidrule(l){9-15} \cmidrule(l){16-22}\cmidrule(l){23-29}
        VGRU-rl  &0.34	&0.47	&0.64	&0.72 &- &- &-   &0.27 &0.40 &0.64 & 0.79 &-&-&- &0.36 & 0.61 &\textbf{0.85} &\textbf{0.92} &-&-&- &0.46 &0.82 &0.95 &1.21 &-&-&-\\
        VGRU-d  &-&-&-&-&-&- &0.1170   &-&-&-&-&-&- &0.1210 &-&-&-&-&-&- &0.0840 &-&-&-&-&-&- &0.1940\\
        TP-RNN  &0.25	&0.41	&0.58	&-	   &0.74	&\textbf{0.77}  &-   	&0.20	&\textbf{0.33}	&\textbf{0.53}	&-	&0.84	&\textbf{1.14} &- &0.26	&0.48	&0.88	&-	&\textbf{0.98}	&1.66 &-  &0.26	&\textbf{0.48}	&0.88	&-	&\textbf{0.98}	&\textbf{1.66} &-\\
        \cmidrule{1-8}\cmidrule(l){9-15} \cmidrule(l){16-22}\cmidrule(l){23-29}
        ERD-QV (ours) &\textbf{0.20} &\textbf{0.34} &\textbf{0.56} &\textbf{0.64} &\textbf{0.72} &0.79 &\textbf{0.0767} &\textbf{0.18} &\textbf{0.33} &\textbf{0.53} &\textbf{0.63} &\textbf{0.78} &1.17 &\textbf{0.0763} &\textbf{0.23} &\textbf{0.47} &0.96 &0.99 &\textbf{0.98} &\textbf{1.58} &\textbf{0.0537}  &\textbf{0.23} &0.59 &\textbf{0.86} &\textbf{0.93} &1.30 &1.75 &\textbf{0.1201}\\
        \cmidrule{1-29}
    \end{tabular}
\end{table*} 
Based on recent advances in motion prediction, we first build a motion prediction network that yields state-of-the-art results. We evaluate our model on the popular motion prediction benchmark that uses the Human 3.6M dataset. We follow the evaluation protocol defined by Jain \etal \shortcite{jain2016structural} that we base on the code from Martinez \etal \shortcite{martinez2017human}. We train the networks for $\num{40500}$ iterations before evaluation. We use the core architecture of Harvey \etal \shortcite{harvey2018recurrent} since their separate encoders allow us to apply our embedding modifiers. In the case of unconstrained prediction however, this is more similar to the Encoder-Recurrent-Decoder (ERD) networks from Fragkiadaki \etal \shortcite{fragkiadaki2015recurrent}. We also apply the velocity-based input representation of Chiu \etal \shortcite{chiu2019action} which seems to be a key component to improve performance. This is empirically shown in our experiments for motion prediction, but as we will see in Section \ref{sec:human_transition}, it doesn't hold for transition generation, where the character state as an input is more informative than velocities to produce correct transitions, evaluated on global angles and positions. 
Another difference lies in our data representation, which is based on quaternions instead of exponential maps. 
We call our architecture for motion prediction ERD-Quaternion Velocity network (ERD-QV). This model is therefore similar to the one depicted in Figure \ref{fig:overview}, with quaternions velocities $\dot{\bq}_t$ as only inputs of the state encoder instead of $\bq_t$ and $\dot{\br}_t$, and without the two other encoders and their inputs. No embedding modifier and no FK are used in this case, and the only loss used is the L1 norm on joint-local quaternions. In this evaluation, the root transform is ignored, to be consistent with previous works.
In Table \ref{table:prediction}, we compare this model with the TP-RNN which obtains to our knowledge state-of-the-art results for Euler angle differences. We also compare with two variants of the VGRU architecture proposed by Gopalakrishnan \etal \shortcite{gopalakrishnan2019neural}, who propose a novel Normalized Power Spectrum Similarity (NPSS) metric for motion prediction that is more correlated to human assessment of quality for motion. Note that in most cases, we improve upon the TP-RNN for angular errors and perform better than the VGRU-d proposed by Gopalakrishnan \etal \shortcite{gopalakrishnan2019neural} on their proposed metric. This allows us to confirm the performance of our chosen architecture as the basis of our transition generation model.  %
\subsection{Walking in-betweens on Human 3.6M}\label{sec:human_transition}
Given our highly performing prediction architecture, we now build upon it to produce a transition generator (TG). We start off by first adding conditioning information about the future target and current offset to the target, and then sequentially add our proposed contributions to show their quantitative benefits on a novel transition benchmark. Even though the Human 3.6M dataset is one of the most used in motion prediction research, most of the actions it contains are ill-suited for long-term prediction or transitions (e.g. \textit{smoking}, \textit{discussion}, \textit{phoning}, ...) as they consists of sporadic, random short movements that are impossible to predict beyond some short time horizons. We thus choose to use a subset of the Human 3.6M dataset consisting only of the three walk-related actions (\textit{walking, walkingdog, walkingtogether}) as they are more interesting to test for transitions over 0.5 seconds long. Like previous studies, we work with a 25Hz sampling rate and thus subsample the original 50Hz data. The walking data subset has $\num{55710}$ frames in total and we keep Subject 5 as the test subject. The test set is composed of windows of motion regularly sampled every 10 frames in the sequences of Subject 5. Our test set thus contains $\num{1419}$ windows with lengths that depend on the evaluation length. In order to evaluate robustness to variable lengths of transitions, we train the models on transitions of random lengths ranging from 0.2 to 2 seconds (5 to 50 frames) and evaluate on lengths going up to 4 seconds (100 frames) to also test generalization to longer time horizons. We train the models for $\num{55000}$ iterations, and report average L2 distances of global quaternions (L2Q) and global positions (L2P):
\begin{align}
    L2Q &= \frac{1}{|\mathcal{D}|}\frac{1}{T} \sum_{s \in \mathcal{D}} \sum_{t=0}^{T-1} \norm{\hat{\bg}_t^s - \bg_t^s}_2\\
    L2P &= \frac{1}{|\mathcal{D}|}\frac{1}{T} \sum_{s \in \mathcal{D}} \sum_{t=0}^{T-1} \norm{\hat{\bp}_t^s - \bp_t^s}_2
\end{align}
where $s$ is a transition sequence of the test set $\mathcal{D}$, and $T$ is the transition length. Note that we compute L2P on normalized global positions using statistics from the training set. Precisely, we extract the global positions' statistics on windows of 70 frames\footnote{We train with transitions of maximum lengths of 50 frames, plus 10 seed frames and 10 frames of future context to visually assess the motion continuation, from which the first frame is the target keyframe. This yields windows of 70 frames in total.} offset by 10 frames in which the motion has been centered around the origin on the horizontal plane. We center the motion by subtracting the mean of the root's XZ positions on all joints' XZ positions. We report the L2P metric as it is arguably a better metric than any angular loss for assessing visual quality of transitions with global displacements. However, it is not complete in that bone orientations might be wrong even with the right positions. We also report NPSS scores, which are based on angular frequency comparisons with the ground truth. Our results are shown in Table \ref{table:transh36}.
\begin{table}[h]
\caption{\textbf{Transition generation benchmark on Human 3.6M}. Models were trained with transition lengths of maximum 50 frames, but are evaluated beyond this horizon, up to 100 frames (4 seconds).}
    \begin{adjustbox}{width=\columnwidth,center}
    \centering
    \footnotesize
    \renewcommand{\arraystretch}{0.85}
    \begin{tabular}{lcccc|cc|c}
        &\multicolumn{6}{c}{\textbf{L2Q}}\\
        \cmidrule(r){2-8}
        Length (frames) &5&10&25&50&75&100&AVG \\
        \midrule 
        Interpolation                                   & \textbf{0.22} &0.43 &0.84 &1.09 &1.48 &2.03 &1.02\\
        TG-QV                                           & 0.36	&0.51 &0.76	&1.08 &1.54	&1.97 &1.04\\
        \cmidrule(lr){1-8}
        TG-Q                                            & 0.33	&0.48 &0.76	&1.05 &1.40	&1.79 &0.97\\
        $ + L_{\mathit{pos}}$                            & 0.32	&0.45 &0.74	&1.04 &1.40	&1.80 &0.97\\
        $ + \bz_{\mathit{tta}}$                          & 0.26	&0.40 &0.70	&0.96 &1.30 &1.67 &0.88\\ 
        $ + \bz_{\mathit{target}}$                       & 0.26	&0.40 &\textbf{0.68} &0.94 &1.22 &1.56 &0.84\\
        $ + L_{\mathit{gen}}$ ($\mathrm{TG}_{\mathit{complete}}$)                            & 0.24	&\textbf{0.38} &\textbf{0.68} &\textbf{0.93} &\textbf{1.20} &\textbf{1.49} &\textbf{0.82}\\
        \\
        &\multicolumn{6}{c}{\textbf{L2P}}\\
        \cmidrule(r){2-8}
        Interpolation                                   & 0.32  &0.69 &1.62 &2.70 &4.34  &6.18 &2.64\\
        TG-QV                                           & 0.48	&0.74 &1.22	&2.44 &4.71 &7.11 &2.78\\
        \cmidrule(lr){1-8}
        TG-Q                                            & 0.50	&0.75 &1.29	&2.27 &4.07	&6.28 &2.53\\
         + $L_{\mathit{pos}}$                           & 0.44	&0.67 &1.22	&2.27 &4.14	&6.39 &2.52\\
        + $\bz_{\mathit{tta}}$                          & 0.34	&0.54 &1.04	&1.82 &3.25	&5.27 &2.04\\ 
        + $\bz_{\mathit{target}}$                       & 0.32	&0.51 &\textbf{0.97}	&\textbf{1.67} &2.85	&4.67 &1.83\\ 
        + $L_{\mathit{gen}}$ ($\mathrm{TG}_{\mathit{complete}}$)                            & \textbf{0.28}	&\textbf{0.48} &\textbf{0.97} &1.68 &\textbf{2.71}	&\textbf{4.37} &\textbf{1.75}\\ 
        \\
        &\multicolumn{6}{c}{\textbf{NPSS}}\\
        \cmidrule(r){2-8}
        Interpolation                                   &\textbf{0.0020} &0.0110 &0.0948 &0.3555 &0.7790 &1.5170 &0.4599\\
        TG-QV                                           &0.0033	&0.0124 &0.0804 &0.3118 &0.8535	&2.0822 &0.5573\\
        \cmidrule(lr){1-8}
        TG-Q                                            &0.0031	&0.0121 &0.0842 &0.3188 &0.8128	&1.8151 &0.5077\\
        + $L_{\mathit{pos}}$                            &0.0031	&0.0117 &0.0811 &0.3105 &0.7988	&1.9658 &0.5285\\
        + $\bz_{\mathit{tta}}$                          &0.0026	&0.0107 &0.0773 &0.2933 &0.7403	&1.6459 &0.4617\\ 
        + $\bz_{\mathit{target}}$                       &0.0026	&0.0107 &\textbf{0.0765} &0.2974 &0.7531 &1.6270 &0.4612\\
        + $L_{\mathit{gen}}$ ($\mathrm{TG}_{\mathit{complete}}$)                           &0.0024	&\textbf{0.0102} &\textbf{0.0765} &\textbf{0.2873} &\textbf{0.6572}	&\textbf{1.3364} &\textbf{0.3950}\\
    \end{tabular}
    \end{adjustbox}
    \label{table:transh36}
\end{table} 
Our first baseline consists of a naive interpolation strategy in which we linearly interpolate the root position and spherically interpolate the quaternions between the keyframes defining the transition. On the very short-term, this is an efficient strategy as motion becomes almost linear in sufficiently small timescales. We then compare transition generators that receive quaternion velocities as input (TG-QV) with one receiving normal quaternions (TG-Q) as depicted in Figure \ref{fig:overview}. Both approaches have similar performance on short transitions, while TG-QV shows worst results on longer transitions. This can be expected with such a model that isn't given a clear representation of the character state at each frame. We thus choose TG-Q as our main baseline onto which we sequentially add our proposed modifications. We first add the global positional loss ($+ L_{\mathit{pos}}$) as an additional training signal, which improves performance on most metrics and lengths. We then add the unconstrained time-to-arrival embedding modifier (+$\bz_{\mathit{tta}}$) and observe our most significant improvement. These effects on 50-frames translations are summarized in Figure \ref{fig:tta_effects}.
%
%
%
Next, we evaluate the effects of our scheduled target embedding modifier $\bz_{\mathit{target}}$. Note that it is turned off for quantitative evaluation. The effects are minor for transitions of 5 and 10 frames, but $\bz_{\mathit{target}}$ is shown to generally improve performances for longer transitions. We argue that these improvements come from the fact that this target noise probably helps generalizing to new sequences as it improves the model's robustness to new or noisy conditioning information. Finally, we obtain our complete model ($\mathrm{TG}_{\mathit{complete}}$) by adding our adversarial loss $L_{\mathit{gen}}$, which interestingly not only improves the visual results of the generated motions, but also most of the quantitative scores.

Qualitatively, enabling the target noise allows the model to produce variations of the same transitions, and it is trivial to control the level of variation by controlling $\sigma_{\mathit{target}}$. We compare our approach to a simpler variant that also aims at inducing stochasticity in the generated transition. In this variant, we aim at potentially disambiguating the missing target information such as velocities by concatenating a random noise vector $\bz_{\mathit{concat}}$ to the target keyframe input $\bq_T$. This is similar to a strategy used in conditional GANs to avoid mode collapse given the condition. Figure \ref{fig:noisecomp} and the accompanying video show typical results obtained with our technique against this more classical technique. 
\subsection{Scaling up with the LaFAN1 dataset}\label{sec:biggerDataset}
Given our model selection based on the Human 3.6M walking benchmark discussed above, we further test our complete model on a novel, high quality motion dataset containing a wide range of actions, often with significant global displacements interesting for in-betweening compared to the Human3.6M dataset. This dataset contains $\num{496672}$ motion frames sampled at 30Hz and captured in a production-grade MOCAP studio. It contains actions performed by 5 subjects, with Subject 5 used as the test set. Similarly to the procedure used for the Human3.6M walking subset, our test set is made of regularly-sampled motion windows. Given the larger size of this dataset we sample our test windows from Subject 5 at every 40 frames, and thus retrieve $\num{2232}$ windows for evaluation. The training statistics for normalization are computed on windows of 50 frames offset by 20 frames. Once again our starting baseline is a normal interpolation. We make public this new dataset along with accompanying code that allows one to extract the same training set and statistics as in this work, to extract the same test set, and to evaluate naive baselines (zero-velocity and interpolation) on this test set for our in-betweening benchmark. We hope this will facilitate future research and comparisons on the task of transition generation.
We train our models on this dataset for $\num{350000}$ iterations on Subjects 1 to 4. We then go on to compare a reconstruction-based, future-conditioned Transition Generator ($\mathrm{TG}_{\mathit{rec}}$) using $L_{\mathit{quat}}$, $L_{\mathit{root}}$, $L_{\mathit{pos}}$ and $L_{\mathit{contacts}}$ with our augmented adversarial Transition Generator ($\mathrm{TG}_{\mathit{complete}}$) that adds our proposed embedding modifiers $\bz_{\mathit{tta}}$, $\bz_{\mathit{tta}}$ and our adversarial loss $L_{\mathit{gen}}$. Results are presented in Table \ref{table:lafanres}.
\begin{table}[h]
    \centering
    \caption{\textbf{Improving in-betweening on the LaFAN1 dataset}. Models were trained with transition lengths of maximum 30 frames (1 second), and are evaluated on 5, 15, 30, and 45 frames.}
    \small
    \begin{tabular}{lccc|c}
        
        &\multicolumn{4}{c}{\textbf{L2Q}}\\
        \cmidrule(r){2-5}
        Length (frames) &5&15&30&45 \\
        \midrule 
        Interpolation           & 0.22 &0.62  &0.98 &1.25\\
        $\mathrm{TG}_{\mathit{rec}}$ & 0.21 &0.48  &0.83 &1.20\\
        $\mathrm{TG}_{\mathit{complete}}$  & \textbf{0.17}	&\textbf{0.42} &\textbf{0.69}	&\textbf{0.94}\\
        &\multicolumn{4}{c}{\textbf{L2P}}\\
        \cmidrule(r){2-5}
        Interpolation           &0.37 &1.25 &2.32 &3.45\\
        $\mathrm{TG}_{\mathit{rec}}$ & 0.32 &0.85  &1.82 &3.89\\
        $\mathrm{TG}_{\mathit{complete}}$   &\textbf{0.23} &\textbf{0.65} &\textbf{1.28} &\textbf{2.24}\\
        &\multicolumn{4}{c}{\textbf{NPSS}}\\
        \cmidrule(r){2-5}
        Interpolation               &0.0023 &0.0391 &0.2013 &0.4493\\
        $\mathrm{TG}_{\mathit{rec}}$     & 0.0025 &0.0304  &0.1608 &0.4547\\
        $\mathrm{TG}_{\mathit{complete}}$   &\textbf{0.0020}	&\textbf{0.0258} &\textbf{0.1328} &\textbf{0.3311}\\
    \end{tabular}
    \label{table:lafanres}
\end{table} 
Our contributions improve performance on all quantitative measurements. On this larger dataset with more complex movements, our proposed in-betweeners surpass interpolation even on the very short transitions, as opposed to what was observed on the Human3.6M walking subset. This motivates the use of our system even on short time-horizons. 
\subsection{Practical use inside an animation software}
\begin{figure}[h]
\begin{center}
\centerline{\includegraphics[width=0.475\textwidth]{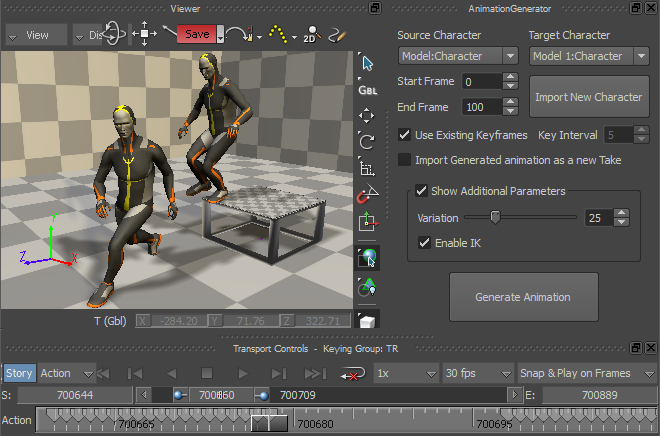}}
  \caption{\textbf{Generating animations inside MotionBuilder.} On the left is a scene where the last seed frame and target keyframe are visible. On the right is our user interface for the plugin that allows, among other things, to specify the level of scheduled target noise for the next generation through the \textit{variation} parameter, and to use the network's contact predictions to apply IK. On the bottom is the timeline where the gap of missing motion is visible.}
  \label{fig:mobu}
\end{center}
\vspace{-0.5cm}
\end{figure}

\begin{figure}[h]
\begin{center}
\centerline{\includegraphics[width=0.475\textwidth]{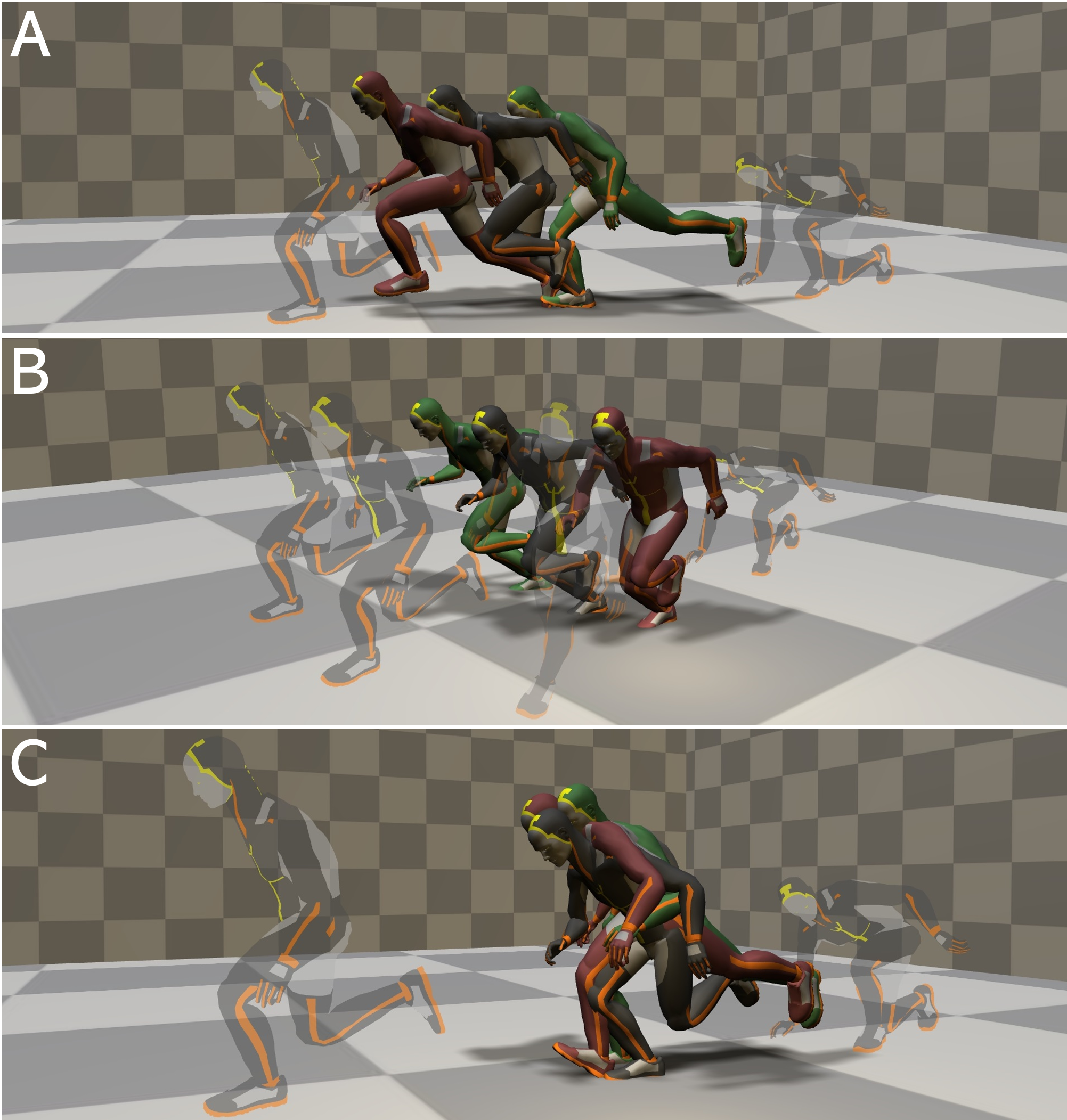}}
  \caption{\textbf{Three types of variations of a \textit{crouch-to-run} transition.} A single frame per generated transition is shown, taken at the same timestep. Semi-transparent poses are the start and end keyframes. \textbf{A}: Temporal variations are obtained by changing the temporal location of the second keyframe. This relies on our time-to-arrival embeddings ($\bz_{\mathit{tta}}$). \textbf{B}: Spatial variations can be obtained by simply moving the target keyframe in space. \textbf{C}: Motion variation are obtained by re-sampling the same transition with our scheduled target noise ($\bz_{\mathit{target}}$) enabled. These results can also be seen in the accompanying video.}
  \label{fig:variations}
\end{center}
\vspace{-0.5cm}
\end{figure}

In order to also qualitatively test our models, we deploy networks trained on LaFAN1 in a custom plugin inside Autodesk's MotionBuilder, a widely used animation authoring and editing software. This enables the use of our model on user-defined keyframes or the generation of transitions between existing clips of animation. Figure \ref{fig:mobu} shows an example scene with an incomplete sequence alongside our user interface for the plugin. 
The \textit{Source Character} is the one from which keyframes are extracted while the generated frames are applied onto the \textit{Target Character}'s skeleton. In this setup it is trivial to re-sample different transitions while controlling the level of target noise through the \textit{Variation} parameter. A variation of 0 makes the model deterministic. Changing the temporal or spatial location of the target keyframes and producing new animations is also trivial. Such examples of variations can be seen in Figure \ref{fig:variations}. The user can decide to apply IK guided by the network's contact predictions through the \textit{Enable IK} checkbox. An example of the workflow and rendered results can be seen in the accompanying video.

\begin{table}[h]
    \centering
    \caption{\textbf{Speed performance summary of our MotionBuilder plugin}. The model inference also includes the IK postprocess. The last column indicates the time taken to produce a string of 10 transitions of 30 frames. Everything is run on a Intel Xeon CPU E5-1650 @ 3.20GHz.}
    \small
    \begin{tabular}{lrrr|r}
        Transition time (s)  & 0.50 &1.00 &2.00 &10 x 1.00\\
        \midrule
         Keyframe extraction (s)& 0.01 &0.01  &0.01 &0.01\\
         Model inference (s)    & 0.30 &0.31  &0.31 &0.40\\
         Applying keyframes (s) & 0.72 &1.05  &1.65 &6.79\\
         Total (s)              & 1.03 &1.37  &1.97 &7.20
    \end{tabular}
    \label{table:perfo}
\end{table} 
The plugin with the loaded neural network takes 170MB of memory. Table \ref{table:perfo} shows a summary of average speed performances of different in-betweening cases. This shows that an animator can use our tool to generate transition candidates almost for free when compared to manually authoring such transitions or finding similar motions in a motion database.

\section{Discussion}
\subsection{Additive modifiers}
We found our time-to-arrival and scheduled target noise additive modifiers to be very effective for robustness to time variations and for enabling sampling capabilities. We explored relatively simpler concatenation-based methods that showed worse performances. We hypothesize that concatenating time-to-arrival or noise dimensions is often less efficient because the neural network can learn to ignore those extra dimensions which are not crucial in the beginning of the training. Additive embedding modifiers however impose a shift in latent space and thus are harder to bypass.  
%
\subsection{Ablated datasets}
In order to gain some insights on the importance of the training set content, we trained two additional models with ablated versions of the LaFAN1 dataset. For the first one, we removed all dance training sequences (approximately 10\% of the data). For the second one, we kept only those sequences, yielding a much smaller dataset (21.5 minutes). Results showed that keeping only the dance sequences yielded similar results as the bigger ablated dataset, but that the full dataset is necessary to generate transitions that stay in a dancing style. This indicates that large amounts of generic data can be as useful as much fewer specialized sequences for a task, but that combining both is key. An example is shown in the accompanying video.

\subsection{Dropping the Triangular-Prism}
When building our motion predictor ERD-QV, we based our input represention on velocities, as suggested with the TP-RNN architecture proposed by \cite{chiu2019action}. However, we did not witness any gains when using their proposed Triangular-Prism RNN (TP-RNN) architecture. Although unclear why, it might be due to the added depth of the network by adding in our case a feed-forward encoder, making the triangular prism architecure unnecessary.

\subsection{Incompleteness of $L_{\mathit{pos}}$}
Although we use FK for our positional loss $L_{\mathit{pos}}$ as suggested by Pavllo \etal \shortcite{pavllo2019modeling}, this loss isn't sufficient to produce fully defined character configurations. Indeed using only this loss may lead to the correct positions of the joints but offers no guarantee for the bone orientations, and led in our experiments to noticeable artifacts especially at the ends of the kinematic chains.

\subsection{Recurrent cell types}
Some recent works on motion prediction prefer Gated Recurrent Units (GRU) over LSTMs for their lower parameter count, but our empirical performance comparisons favored LSTMs over GRUs.

\section{Limitations and future work}
A more informative way of representing the current offset to the target $\bo_t$ would be to include positional-offsets in the representation. For this to be informative however, it would need to rely on character-local or global positions, which require FK. Although it is possible to perform FK inside the network at every step of generation, the backward pass during training becomes prohibitively slow justifying our use of root offset and rotational offsets only. 

As with many data-driven approaches, our method struggles to generate transitions for which conditions are unrealistic, or outside the range covered by the training set.

Our scheduled target noise allows us to modify to some extent the manner in which a character reaches its target, reminiscent of changing the style of the motion, but doesn't allow yet to have control over those variations. Style control given a fixed context would be very interesting but is out of scope of this work.


\section{Conclusion}
In this work we first showed that state-of-the-art motion predictors cannot be converted into robust transition generators by simply adding conditioning information about the target keyframe. We proposed a time-to-arrival embedding modifier to allow robustness to transition lengths, and a scheduled target noise modifier to allow robustness to target keyframe variations and to enable sampling capabilities in the system. We showed how such a system allows animators to quickly generate quality motion between sparse keyframes inside an animation software. We also presented LaFAN1, a new high quality dataset well suited for transition generation benchmarking.

\begin{acks}
We thank Ubisoft Montreal, the Natural Sciences and Engineering Research Council of Canada and Mitacs for their support. We also thank Daniel Holden, Julien Roy, Paul Barde, Marc-André Carbonneau and Olivier Pomarez for their support and valuable feedback.
\end{acks}


\bibliographystyle{ACM-Reference-Format}
\bibliography{references}

\appendix
\section{Appendix}
\subsection{Sliding critics}\label{app:critics}
\vskip -0.4cm
\begin{figure}[H]
\begin{center}
\centerline{\includegraphics[width=0.35\textwidth]{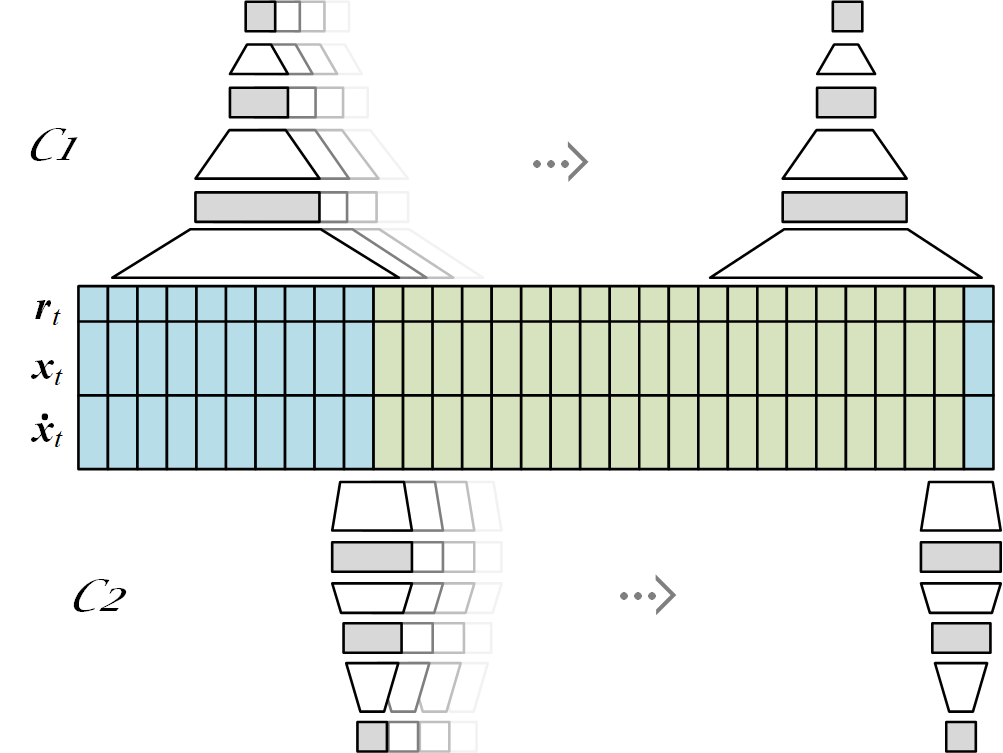}}
  \caption{\textbf{Visual summary of the two timescales critics.} Blue frames are the given contexts and green frames correspond to the transition. First and last critic positions are shown without transparency. At the beginning and end of transitions, the critics are conditional in that they include ground-truth context in their input sequences. Scalar scores at each timestep are averaged to get the final score.}
  \label{fig:critics}
\end{center}
\vspace{-0.5cm}
\end{figure}

\subsection{Time-to-arrival embedding visualization}\label{app:tta}
\vskip -0.4cm
\begin{figure}[H]
\begin{center}
\centerline{\includegraphics[width=0.38\textwidth]{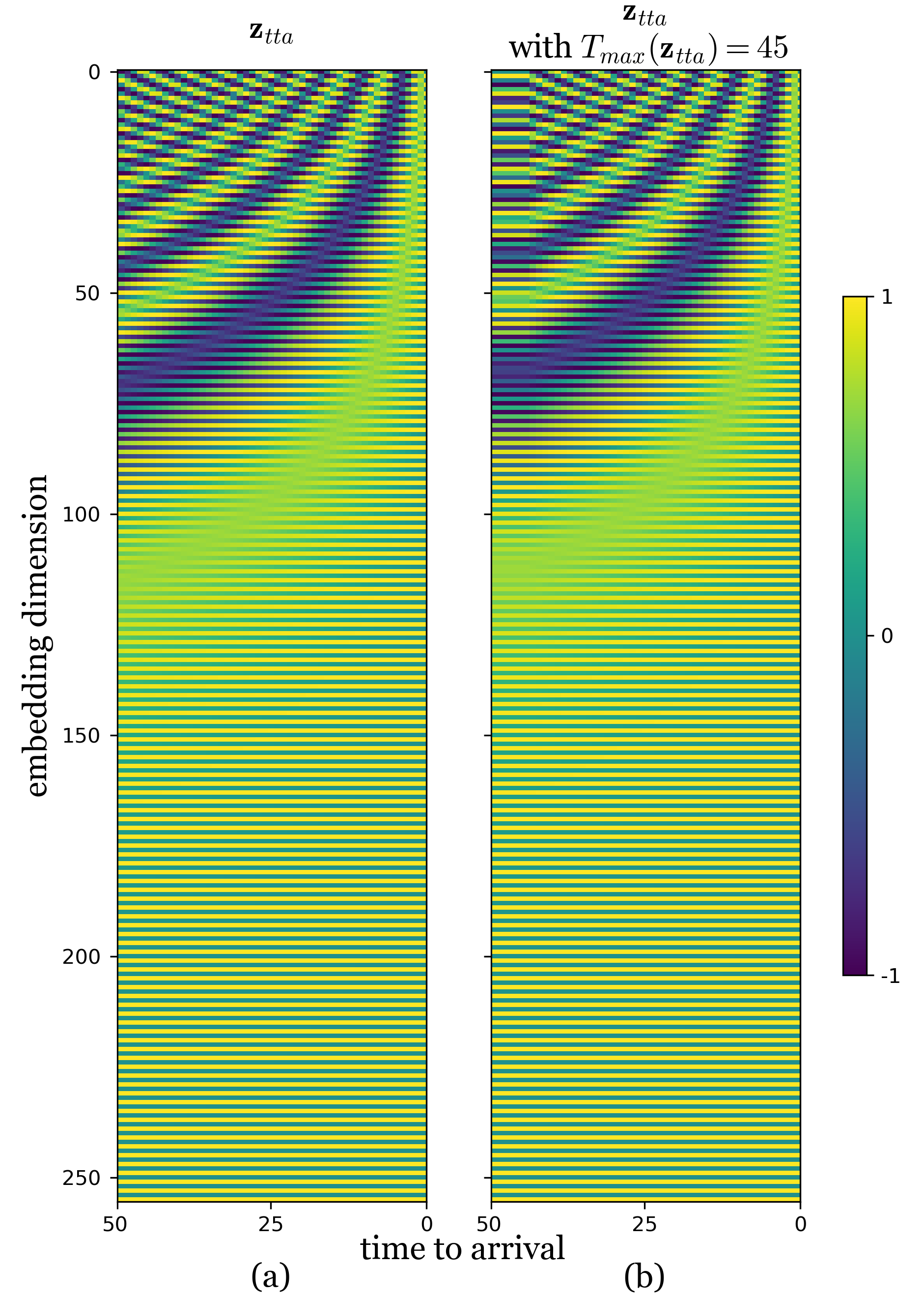}}
  \caption{\textbf{Visual depiction of time-to-arrival embeddings.} Sub-figure (b) shows the effect of using $T_{\mathit{max}}(\bz_{\mathit{tta}})$, which in practice improves performances when generalizing to longer transitions as it prevents initializing the LSTM hidden state with novel embeddings.}
  \label{fig:tta}
\end{center}
\vspace{-0.5cm}
\end{figure}

\end{document}